\documentclass[lettersize,journal]{IEEEtran}

\usepackage{amsmath,amsfonts}
\usepackage{algorithmic}
\usepackage{algorithm}
\usepackage{array}
\usepackage{subfig}
\usepackage{textcomp}
\usepackage{stfloats}
\usepackage{url}
\usepackage{verbatim}
\usepackage{graphicx}
\usepackage{cite}
\newcommand{\reffig}[1]{Fig.~\ref{#1}}
\newcommand{\reftab}[1]{Table~\ref{#1}}
\newcommand{\refsec}[1]{Section~\ref{#1}}

\begin{document}

\title{YOLO-Former: YOLO Shakes Hand With ViT}

\author{Javad Khoramdel \and Ahmad Moori \and Yasamin Borhani \and Armin Ghanbarzadeh \and Esmaeil Najafi \\
Tarbiat Modares University \and Faculty of Mechanical Engineering, K. N. Toosi University of Technology \\
Tehran, Iran \\
{\tt\small j.khoramdel@modares.ac.ir} \and {\tt\small ahmadmoori@email.kntu.ac.ir} \and {\tt\small borhaniyasamin@gmail.com} \and {\tt\small agz1986@gmail.com} \and {\tt\small najafi.e@kntu.ac.ir}}



\maketitle

\begin{abstract}
The proposed YOLO-Former method seamlessly integrates the ideas of transformer and YOLOv4 to create a highly accurate and efficient object detection system. The method leverages the fast inference speed of YOLOv4 and incorporates the advantages of the transformer architecture through the integration of convolutional attention and transformer modules. The results demonstrate the effectiveness of the proposed approach, with a mean average precision (mAP) of 85.76\% on the Pascal VOC dataset, while maintaining high prediction speed with a frame rate of 10.85 frames per second. The contribution of this work lies in the demonstration of how the innovative combination of these two state-of-the-art techniques can lead to further improvements in the field of object detection.
\end{abstract}


\section{Introduction}
Many computer vision tasks, such as image classification, image segmentation, and object detection, are dominated by deep neural networks due to the recent advancements in deep learning. Object detection is the task of detecting instances of semantic objects of a certain class in digital images and videos \cite{papageorgiou1998general}. Some applications of such systems are license plate character recognition, object tracking, human face and body detection and recognition, activity recognition, medical imaging, advanced driving assistant systems, manufacturing industry, and robotics. 

With the advent of big-data and higher processing power, the deep neural network based methods for object detection have become more popular. These networks are capable of end-to-end object detection without the need of additional components and are mostly based on convolutional neural networks \cite{lecun1995convolutional}. 
The state-of-the-art object detection methods can be further categorized into two main categories. First, region proposal based models that prioritize detection accuracy over inference speed such as RCNN~\cite{girshick2014rich}, fast RCNN~\cite{girshick2015fast}, mask RCNN \cite{he2017mask}. Second, one-stage detection models that have high inference speeds and are capable of achieving real time detection. The examples of one-stage models include single shot multibox detector~(SSD)~\cite{liu2016ssd}, you only look once (YOLO)~\cite{redmon2018yolov3}, EfficientDet~\cite{tan2020efficientdet}, RetinaNet~\cite{lin2017focal}, CenterNet~\cite{duan2019centernet}, and HourGlass~\cite{newell2016stacked}. 

Although all the previously mentioned object detectors rely solely on the convolutional and pooling layers, the impressive results of Vision Transformer (ViT) \cite{dosovitskiyimage} which is based on attention layers, has inspired ViT-YOLO \cite{zhang2021vit}, and DETR \cite{carion2020end} to develop object detectors based on the idea of the transformer. The detection transformer (DETR) framework uses the transformer encoder-decoder-based architecture to perform end-to-end object detection~\cite{carion2020end}. The ViT-YOLO embeds the scaled dot multi-head attention layer at the end of the YOLOv4 backbone by flattening the feature maps before the attention layer. It then reshapes the attention layer outputs to 2D to be consistent with the remainder of the network \cite{zhang2021vit}. 

This paper improves the accuracy of YOLOv4 by introducing the YOLO-Former algorithm that employs a novel convolutional self-attention module (CSAM) in the YOLOv4 structure. The CSAM is developed based on the scaled dot self-attention (SDSA). In order to connect the proposed CSAM to other components in the network, a convolutional transformer module has been implemented. The presented object detector is further enhanced by using several augmentation policies to increase its generalization capability. As such, the YOLO-Former provides more accurate results on the Pascal VOC dataset, while preserving the real-time execution property.

The structure of the paper is as follows. \refsec{sec:Background} presents a summary of the studies on augmentations, regularization, and attention mechanisms. The network structure and developed modules are discussed in \refsec{sec:Network Structure}. A detailed description of the experiments conducted with the implemented model and the YOLOv4 dataset, training configuration, and evaluation is available in \refsec{sec:experiment}. The results and comparison to the literature are discussed in \refsec{sec:Results and Discussion} and, finally, \refsec{sec:conclusion} concludes the paper.

\section{Background}
\label{sec:Background}

A brief review of the formerly conducted studies on augmentations, regularization, and attention mechanisms is given in the following. 

\subsection{Augmentation}
\label{sec:aug}


The great impact of the augmentation on extending the generalization ability of the models has made it inseparable from image processing. A network can benefit from augmentation methods such as translation, color jittering, rotation, etc. not only as a means of providing more data, but also as a means of making it less sensitive to these transformations \cite{shorten2019survey}. For instance, occlusion is a challenge problem in image recognition. One solution for that is introduced as the cutout method, which makes the used dataset more versatile \cite{devries2017improved}. In this technique, a random region of images is covered by a rectangle which its size can be chosen according to the size of the objects in each image. Thus, the network should not learn features that rely on the whole object of interest. The same inspiration underlies other methods, such as GridMask \cite{chen2020gridmask} and Hide-and-seek (HaS)\cite{singh2018hide}.


Augmenting across a batch of samples can be beneficial as it extends the vicinity of the dataset as multiple instances of multiple images are mixed to produce a new picture \cite{zhang2017mixup}. Various augmentations in image classification are applied to a batch of images like mixup \cite{zhang2017mixup}, cutmix \cite{yun2019cutmix}, and puzzle mix \cite{kim2020puzzle}. Although attempts have been made to extend the applicability of these techniques beyond image classification to object detection \cite{zhang2017mixup}, a very compelling method called mosaic augmentation is implemented in \cite{bochkovskiy2020yolov4}.



Studies suggest that the severity and number of augmentation techniques used during the training affect the model accuracy \cite{cubuk2019autoaugment, cubuk2020randaugment, hendrycks2019AugMix}. By training a reinforcement learning agent on a small dataset, AutoAugment attempts to find an augmentation policy for combining the augmentation transformations. The high computational cost of AutoAugment encouraged the authors of \cite{cubuk2020randaugment} to develop RandAugment, which parameterizes the data augmentation process with only two parameters; the number of operations (N), and severity (M). Combining RandAugment \cite{cubuk2020randaugment} and mixup \cite{zhang2017mixup}, AugMix augments an image separately in different chains. An augmented image is created by weighted summation of augmentation chains using coefficients from a Dirichlet distribution. Finally, the coefficients from a Beta distribution are drawn to calculate the weighted sum of the original and augmented images.

\subsection{Regularization}
\label{sec:reg}
As with augmentation, the overfitting problem can be reduced by regularization techniques like dropout. Although dropout works well with fully connected layers, the authors of \cite{ghiasi2018dropblock} have developed Dropblock as a method for convolutional layers. Rather than randomly dropping features at random locations, Dropblock drops a connected region. According to their study, decreasing the probability of keeping blocks is more effective than utilizing a fixed probability.

Some methods are only applicable to a specific structure, like shake-shake regularization \cite{gastaldi2017shake}, which can be applied to a multi-branch network. This approach is developed for a three-branch network; two branches are multiplied by small random numbers, then summed up with the third branch during the training forward pass, and different random numbers from a Beta distribution are used as multipliers during backpropagation. These two branches are multiplied by 0.5 at test time.
\subsection{Attention Mechanisms}
\label{sec:attention}
The application of attention mechanisms in the artificial neural network has been associated with NLP tasks \cite{sutskever2014sequence}. In machine translation, the network should concentrate on certain parts of the input sequence from the source language to predict a word in the target language. An attention mechanism is proposed in \cite{bahdanau2014neural} that could help the network to pay attention. This work encouraged other researchers to investigate the applicability of attention for solving different tasks \cite{wu2016google, luong2015effective, yang2016hierarchical, hermann2015teaching}. Formerly, the common choices for solving the NLP tasks like machine translation were recurrent and convolutional neural networks. The authors of \cite{vaswani2017attention} suggested a different architecture called Transformer. Unlike the other works which combined attention with either recurrent or convolutional neural networks, Transformer was only based on the  scaled dot multi-head self-attention (SDMHSA). They claimed that attention could solve the machine translation task on its own. This concept has been investigated in other studies like BERT \cite{devlinbert}.

The promising results of using attention in the NLP field has motivated computer vision scientists to improve their results by adding attention to their networks \cite{wang2017residual,chen2016attention, hu2018squeeze, zhao2016deep}. In \cite{woo2018cbam}, convolutional block attention module (CBAM) has been introduced for convolutional neural networks. This module includs two sub-modules. A spatial attention module (SAM) and a channel attention module (CAM). The authors embedded the CBAM in the structure of a number of the state-of-the-art architectures like ResNet50 \cite{he2016deep}, ResNeXt50 \cite{xie2017aggregated}, and MobileNet \cite{howard2017mobilenets}. By taking advantage of attention, they have achieved higher accuracy in image classification on ImageNet \cite{deng2009imagenet} and object detection on Pascal VOC \cite{everingham2015pascal} and Microsoft COCO \cite{lin2014microsoft}. The vision transformer (ViT) has bridged the gap between image classification and transformer architecture by dealing with an image as a sequence of patches. This network has achieved state-of-the-art accuracy on ImageNet classification. Similar to \cite{devlinbert} and \cite{vaswani2017attention}, ViT only uses the MHSDSA as the main component all over the network \cite{dosovitskiyimage}.


\section{Network Structure}
\label{sec:Network Structure}
The YOLOv4 architecture can be divided into three sub-networks: the backbone, the neck, and the head. The backbone of YOLOv4 is called CPS-Darknet-53. The CPS-Darknet-53 extracts feature from the input image and generate output at three levels. The first level output has the highest spatial resolution and is suitable for detecting small-sized objects. The second level output has less spatial resolution than the first, making it appropriate for finding medium-sized objects in the image. The feature map has more depth than the first stage feature map at this stage. The third and last stage output has the deepest feature map with the least spatial resolution. The YOLOv4 neck takes these feature maps and up-samples the lowest resolution feature map with the bi-linear interpolation method to match the spatial resolution of the second stage feature map. Then this up-sampled feature map is then concatenated with the second-level feature map to help the mid-level resolution feature map enrich the features for detecting medium-sized objects. The obtained feature map is up-sampled and concatenated with the highest-resolution feature map. The YOLOv4 head receives the feature maps from the neck to detect objects at three scales.

It is evident that residual blocks play a vital role inside the YOLOv4 backbone due to 23 residual blocks in CPS-Darknet-53. Motivated by the ViT transformer block, a transformer attention block is implemented and utilized to replace the residual blocks in CPS-Darknet-53. Replacement of the residual blocks is decided due to the fact that the attention block contains residual connections, and a network can learn to bypass the attention mechanism using the residual connections if necessary. Consequently, not only is the residual property preserved, but the network can also learn to pay attention to the areas of interest. The convolutional transformer and the convolutional self-attention modules will be explained in \refsec{sec:tb} and \refsec{sec:cam} respectively.

\subsection{Convolutional Transformer Module}
\label{sec:tb}
\begin{figure}[!t]
\centering
\includegraphics[width=\columnwidth]{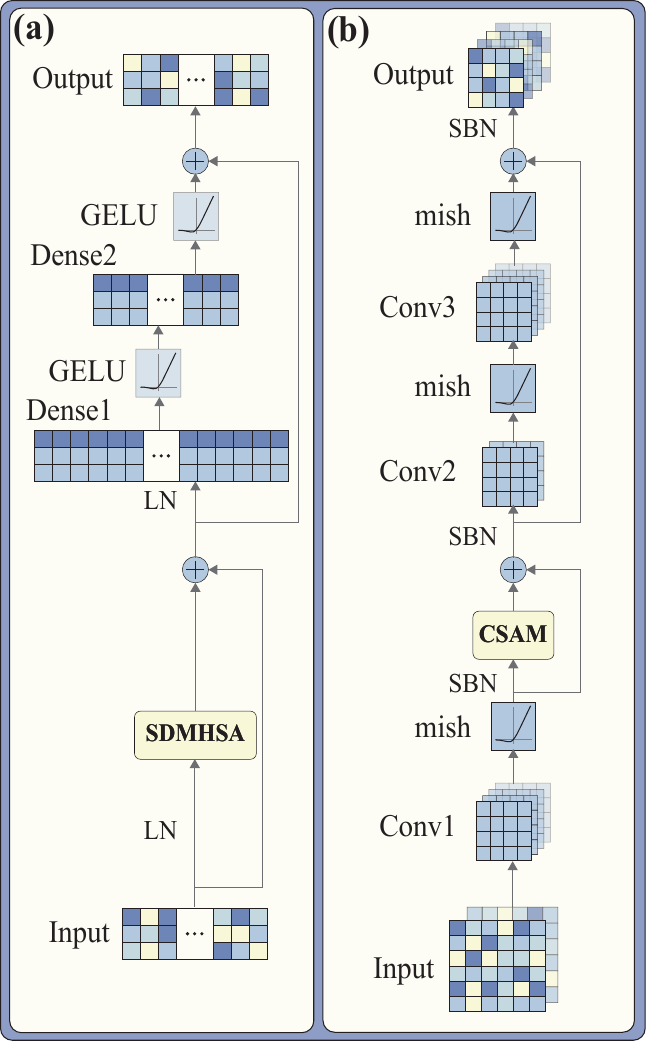}  
\caption{The transformer layer structure for (a) Yolov4 and (b) YOLO-Former. Each module initially pre-processes the input features and feeds it into the specialized attention module. The sum of attention layer output and the input features are processed differently to obtain the output.}\label{fig:transformerlayer}
\end{figure}

Like residual blocks in ResNet architectures, transformer layers are building blocks in ViT models. The block diagram for these layers is demonstrated in \reffig{fig:transformerlayer}(a). Before applying the scaled dot multi-head self-attention (SDMHSA), layer normalization (LN) normalizes the features. A residual connection adds the initial features to the output of the attention layer. Afterward, the results are normalized once more by layer normalization. Then, the normalized features are given to two consecutive dense layers (Dense1 and Dense2) with GeLU activation. Another residual connection combines the output of the SDMHSA with the output of the dense layers.

Inspired by this layer, a convolutional transformer layer is implemented, shown in \reffig{fig:transformerlayer}(b). While retaining the same overall structure as the transformer layer, minor changes have been made to make the module consistent with the convolutional network structure. The features' dimensions remain unchanged in ViT models; however, the depth of feature maps usually changes in CNNs. Accordingly, to ensure the dimensions match inside the residual connections, Conv1 is added to the module. Also, the SDMHSA, dense layers are replaced with the convolutiona self-attention module (CSAM) and convolutional layers (Conv2 and Conv3), respectively. Since batch normalization outperforms layer normalization in CNNs, synchronized batch normalization (SBN) is preferred.

\subsection{Convolutional self-attention Module}
\label{sec:cam}

\begin{figure*}[!t]
	\centering
	\includegraphics[width=.97\textwidth]{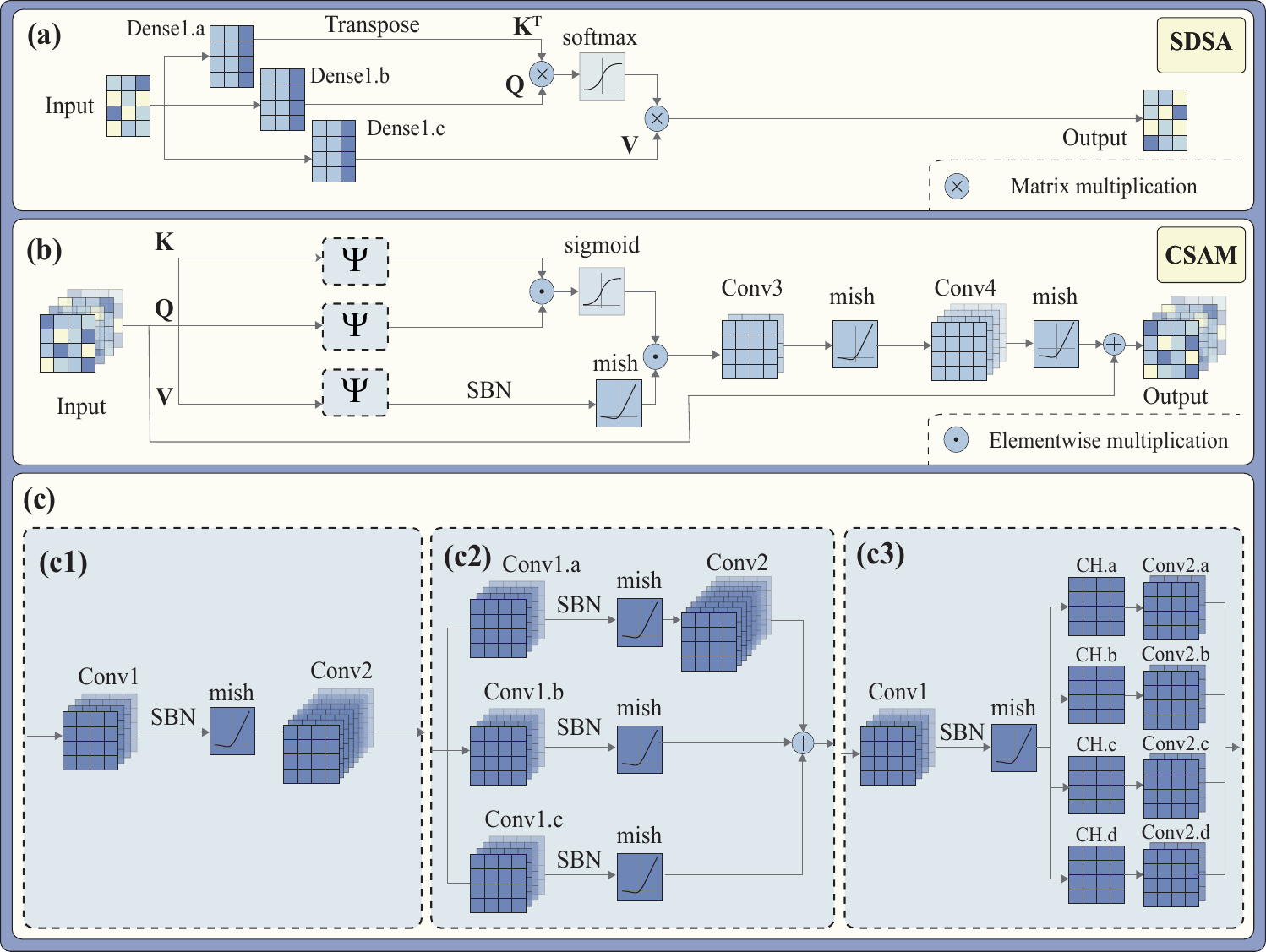}  
	\caption{Attention modules used in (a) Yolov4 and (b) YOLO-Former. Each input is divided to three branches: Key, Query, and Value. These branches are then processed and multiplied according to the multiplication convention stated in each algorithm. They are then processed in a final stage and output (They then go through one final stage of processing before being output). (c) Potential building blocks used in the YOLO-Former attention module for the model iterations described in the paper: single-head (c1), multi-branch (c2), and multi-head (c3)}\label{fig:attblocks}
\end{figure*}

A fundamental component of ViT is the scaled dot self-attention (SDSA) mechanism. As shown in \reffig{fig:attblocks}(a), the linear projection of the input forms the Query (Q), Key (K), and Value (V). Attention scores are calculated by taking softmax from the matrix multiplication of Q and K\textsuperscript{T}. The final output results from a matrix multiplication between the value and the attention scores. Multi-head self-attention is created by feeding the same input to multiple SDSAs and concatenating the outputs.

Based on the idea of SDSA, a convolutional self-attention mechanism has been developed in this paper, shown in \reffig{fig:attblocks}(b). Same as in the ViT, the same inputs are given to the Query, Key, and Value gates. Each gate processes the input by a series of operations shown with $\Psi$ in \reffig{fig:attblocks}(b). The $\Psi$ operations are similar across all three gates.

A variety of attention modules are formed depending on what occurs within the $\Psi$. In the simplest scheme, a 1x1 convolution (Conv1) followed by the synchronized batch-~normalization (SBN) and the mish activation, then; convolved with the 3x3 kernels (Conv2) which creates the output (single-head attention, \reffig{fig:attblocks}(c1)). With the aim of finding an analogous to SDMHSA in convolutional configurations, other alternatives are also developed. A variant is implemented by processing the input in three parallel branches and adding the outputs together (multi-branch attention, \reffig{fig:attblocks}(c2)). There are two branches with only a 1x1 convolution (Conv1.b and Conv1.c), SBN, and mish activation, and one branch with an additional 3x3 convolution (Conv2). The next form divides the mish block result by its depth into four feature maps (CH.a, CH.b, CH.c, and CH.d) and processes each of them separately using 3x3 convolutions and then, concatenates them (multi-head attention, see \reffig{fig:attblocks}(c3)). Multi-head and multi-branch concepts are combined by integrating the multi-head concept into each branch of the multi-branch attention module (multi-head multi-branch attention).

As \reffig{fig:attblocks}(b) indicates, no matter which variant is applied, the activation map is obtained via the multiplication of Query and Key, activated by the sigmoid function. The $\Psi$ output in the Value gates goes through SBN layer and mish activation. Afterward, the Value s multiplied by the attention map. The result of this operation is convolved with a 1x1 kernel (Conv3), and like the other 1x1 convolutions, the output is fed to mish function accordingly. The same procedure is repeated with a convolutional layer with a 3x3 kernel (Conv4) to acquire the outcome of the attention branch. The final output of the attention mechanism is obtained by summing up its branch output with the residual connection from the input so that the network can learn to bypass the attention if required. In contrast with SDSA, all the multiplications are element-wise in the proposed module.

\section{Experiment}
\label{sec:experiment}
This section outlines the details of the conducted experiments, including the dataset, the augmentation techniques, hyperparameters, the training and evaluation procedures.


\subsection{Dataset}
This study focuses on the Pascal VOC as the target dataset. This dataset has 21,503 images (16,551 for the training set and 4952 for the test set), including bounding box annotations for the instances of 20 different categories. Since the number of training samples has not been adequate to overcome the overfitting, 100,000 additional images from the Microsoft COCO dataset have been added to the training set. The Microsoft COCO dataset has labels for 80 categories, 20 of which are the same as Pascal VOC. In order to merge these two datasets, common categories are preserved, and the other classes are neglected.

\begin{figure*}[!t]
	\centering
	\subfloat[(a)]{
		\centering
		\includegraphics[width=.9\linewidth]{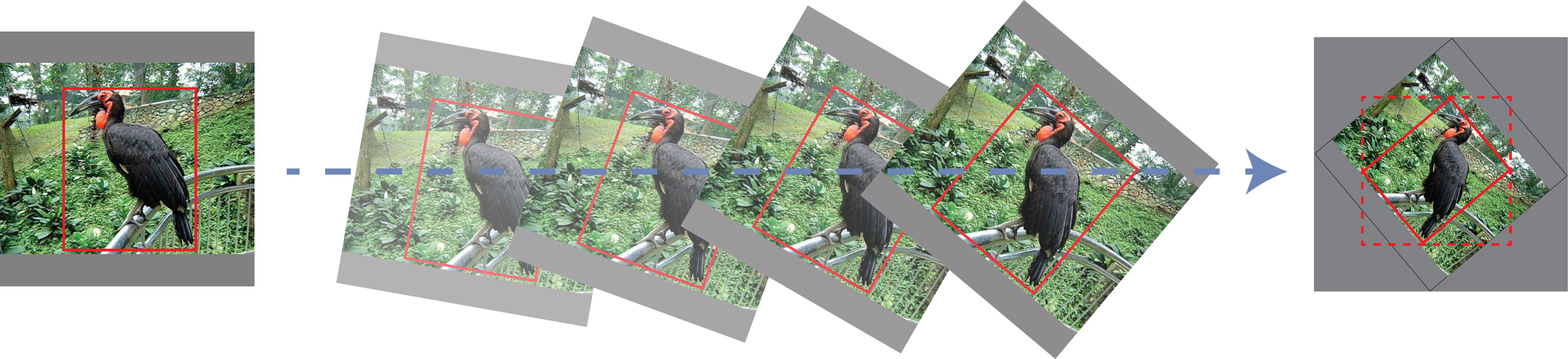}
		\label{fig:rot}
	}
	\\
	\vspace{0.5cm}
	\subfloat[(b)]{
		\centering
		\includegraphics[width=.9\linewidth]{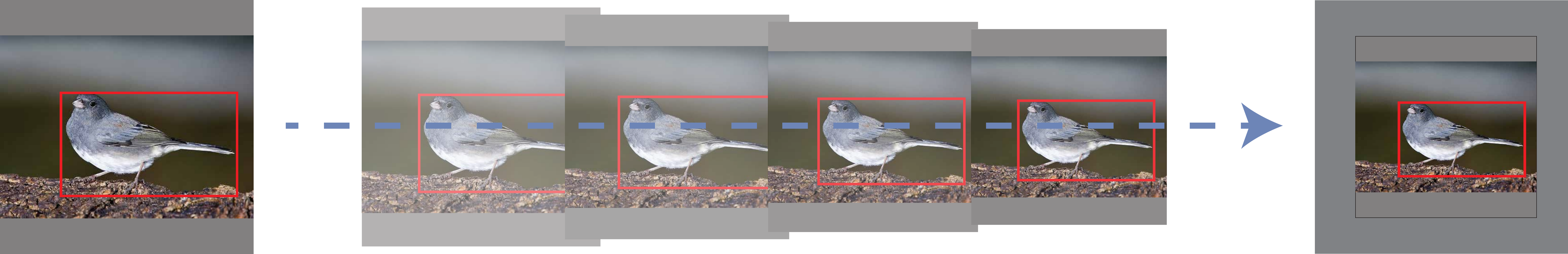}
		\label{fig:zoom}
	}
	\caption{The process of two augmentations (a) constrained rotation and (b) zoom out, being applied to sample images from the dataset and maintain all parts of the image as well as the original size.}
	\label{fig:aug}
\end{figure*}

\subsection{Implemented Augmentations}

Several augmentations are used in this research; one of them is rotation. Although rotation is a popular augmentation technique for image classification, naively rotating the image in object detection might cause objects of interest to slip-outside the image boundaries. In addition, prior bounding boxes are no longer credible. Hence, the image is padded so that the corners of the unrotated image lie on the boundaries of the rotated image. Then, the image is resized to match the initial resolution, as is shown in \reffig{fig:aug}(a). In order to accommodate for homogeneous transformation, the bounding boxes are also rotated along the center of the image. Finally, the smallest rectangle, which includes the rotated bounding box, is selected as the new bounding box.

Another augmentation comes from the fact that it is conventional to train fully convolutional neural networks with variable input image resolutions to make them scale invariant \cite{he2015spatial, lowe2004distinctive}. For example, YOLOv4 is trained with varying image sizes during the training process in the darknet framework. In every ten iterations, nine are trained with the basic resolution, and the remaining one is trained with a random resolution, higher or lower than the prior size. Currently, TPUs are not capable of handling variable tensors during training. Instead of using a variety of resolutions, we use zoom-out augmentation as shown in  \reffig{fig:aug}(b). The input image is randomly resized to the lower resolutions. Then it is padded to maintain the original size of the picture.

Besides constrained rotation and zoom-out, other operations are utilized, including color jittering, translation, cropping, horizontal flip, posterizing, cut-out, solarizing, inversion, sharpening, and mosaic augmentation. These operations can be divided into two groups: (i) geometrical operations (e.g., translation, crop, zoom-out, flip, constrained rotation) and (ii) non-geometric operations (e.g., cut-out, color jittering, posterizing, etc.). Geometrical methods are successively applied in random order with a probability of 50\% for each operation. Non-geometric methods are combined with either RandAugment or AugMix. RandAugment's number of augmentation layers (N) and magnitude of operations (M) are set to 2 and 10, respectively. Hyper-parameters for AugMix include three chains of augmentation with a severity of 7 for each technique. The depth of each chain is randomly selected between 1 and 3. The coefficients for mixing the chains were drawn from a beta distribution with $\alpha = \beta = 1$. By using mosaic augmentation smaller batch sizes can be utilized. Moreover, offline mosaic augmentation needs less memory, hence storage problems can be avoided. 

\subsection{Implemented Regularizations}
The implemented regularization techniques are scheduled drop block (SDB) \cite{ghiasi2018dropblock}, shake-shake \cite{gastaldi2017shake}, and L2. In the experiments utilized with SDB, only the last three layers before the YOLO layers are regularized with 3x3 blocks. Starting from the probability of 1 for keeping the blocks, this rate is scheduled to achieve the probability of 0.90 at the last epoch. The shake-shake regularization are only applied to models that have multi-branch attention arrangements, as the nature of this regularization requires it. In the presence of shake-shake the two branches with only 1x1 convolutions (Conv1.b and Conv1.c in \reffig{fig:attblocks}) are multiplied by random numbers in feedforward and backpropagation during the training. For the L2, the coefficient is set to 0.0005.

\begin{table*}[!t]
	\begin{center}
		\begin{tabular}{|c|l|c|c|l|c|c|}
			\hline
			
			{}&Model & Input Res. & Aug. Policy & Regularization &  Attention Module & mAP \%\\
			\hline
			1 & YOLOv4 & 416x416 & AugMix &  L2 & -&{83.27}\\
			2 & YOLOv4 & 512x512 & AugMix &  L2 &-& {83.75}\\
			3 & YOLOv4 & 512x512 & RandAug. &  L2 &-& {85.21}\\
			4 & YOLOv4 & 512x512 & RandAug. &  L2 \& SDB& -&{85.06}\\
			5 & YOLO-Former & 416x416 & AugMix &  L2 &single-head &{83.32}\\
			6 & YOLO-Former & 512x512 & AugMix&  L2 & single-head &{84.26}\\
			7 & YOLO-Former & 512x512 & RandAug. &  L2 & single-head &85.76\\
			8 & YOLO-Former & 512x512 & RandAug. &  L2 \& SDB & single-head & {85.37}\\
			9 & YOLO-Former & 512x512 & RandAug. &  L2   & multi-branch & {85.40}\\
			10 & YOLO-Former & 512x512 & RandAug. &  L2 \& shake-shake & multi-branch & {85.60}\\
			11 & YOLO-Former & 512x512 & RandAug. &  L2  & multi-head & {84.63}\\
			12 & YOLO-Former & 512x512 & RandAug. &  L2 \& shake-shake & multi-head multi-branch& {\textbf{86.01}}\\
			\hline
		\end{tabular}
	\end{center}
	
	\caption{The mean average precision (mAP) of YOLO-Former and YOLOv4 with different augmentation techniques, input resolutions, and regularization methods.}
	\label{tab:res1}
\end{table*}

\begin{figure}[!t]
	\centering
	\includegraphics[width=\linewidth]{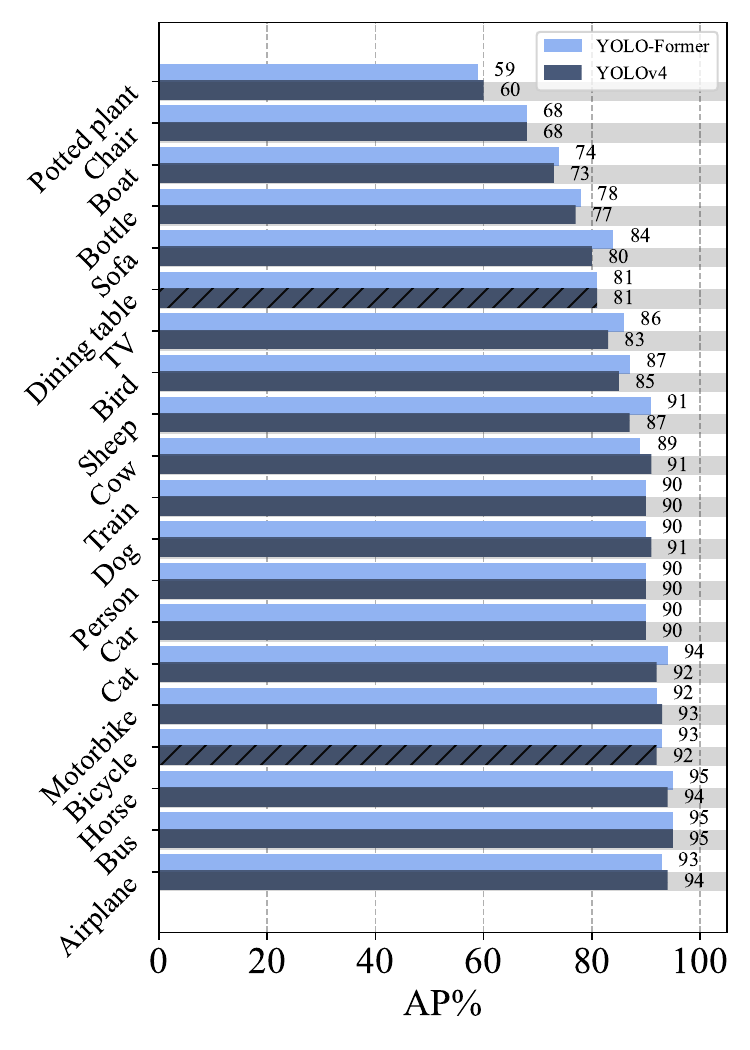}
	\centering
	\caption{Comparison between YOLOv4 and YOLO-Former's average precision (AP) on Pascal VOC classes.}
	\label{fig:res1}
\end{figure}

\subsection{Training and Evaluation}
In all the experiments, the models are trained on TPU with 8 replicas with a mini-batch size of 32 (global batch size of 256) for 225 epochs. The objective function is the summation of the GIoU \cite{rezatofighi2019generalized}, the focal loss \cite{linfocal}, and the binary cross-entropy for localization, foreground-background identification, and classification. The labels for classification are smoothened with a factor of 0.01. To avoid the instability at the beginning, the learning rate increases linearly from 0 to 0.0026 in the first 20 epochs; then cosine decay learning rate scheduler \cite{loshchilov2016sgdr} is utilized to decrease the learning rate down to 0 at the last iteration. All the batch normalization layers in the networks are synchronized. SGD optimizes the model weights with a momentum of 0.996. At the end of each epoch, the mAP is calculated on the test set of Pascal VOC with an IOU threshold of 50\%. The best obtained mAP is reported as the final result at each trial.

\section{Results and Discussion}
\label{sec:Results and Discussion}
The primitive trials are accomplished with YOLO-Former, based on the single-head attention module. Initially, no augmentations or additional images are used for training the YOLO-Former model, but the results have been inferior (39.76\% mAP). Adding simple augmentation techniques such as random translation, random crop, and changing the image's hue, contrast, and brightness improves the results up to 47.24\%. Further improvement have been obtained by adding more augmentations like mosaic augmentation, constrained rotation, and zoom-out (64.09\%). Depite the test accuracy, the training accuracy is considerably high (98\% mAP) in all the immediate experiments. This considerable gap motivates to extend the experiments with a more extensive training set.

After the immediate trials, YOLOv4 and YOLO-Former are trained on the combination of COCO and Pascal VOC training sets. A breakdown of the experiment settings and results is presented in \reftab{tab:res1}. Starting with the input resolution of 416, these models are trained with AugMix as the policy for augmentation and L2 as the regularization. The YOLO-Former is able to achieve an mAP of 83.32\% which is slightly better than YOLO (83.27\%). This vast improvement has been made thanks to the additional data from the COCO dataset. With the same setting, increasing the input resolution to 512 helps the YOLO-Former to enhance the mAP up to 84.26\%. Training time also increases from 100 to 144 hours when the input resolution is increased. The same experiment is executed by switching the augmentation policy from AugMix to RandAugment. YOLOv4 and YOLO-Former achieves better results with the RandAugment policy than AugMix. In order to investigate the possibility of achieving better, scheduled DropBlock regularization is added to the networks in the last experiment as the regularization procedure. Still, it causes a drop in the mAP for both models. 

Aside from achieving more accurate results than YOLOv4, other variations of the proposed attention module are investigated for the possibility of achieving better results than the single-head (SH) configuration (85.76\%). The network based on the multi-branch (MB) attention module configuration is trained with and without shake-shake regularization. While both obtained mAPs are less than single-head configuration, the shake-shake regularization increases the multi-branch variant's mAP from 85.42\% to 85.60\%. The multi-head design provided better results than YOLOv4, but less than the single-head design (85.37\%). The multi-head multi-branch (MHMB) attention module regularized by shake-shake helps the YOLO-Former to gain the highest mAP (86.01\%).

\begin{figure}[t]
	\centering
	\includegraphics[width=\columnwidth]{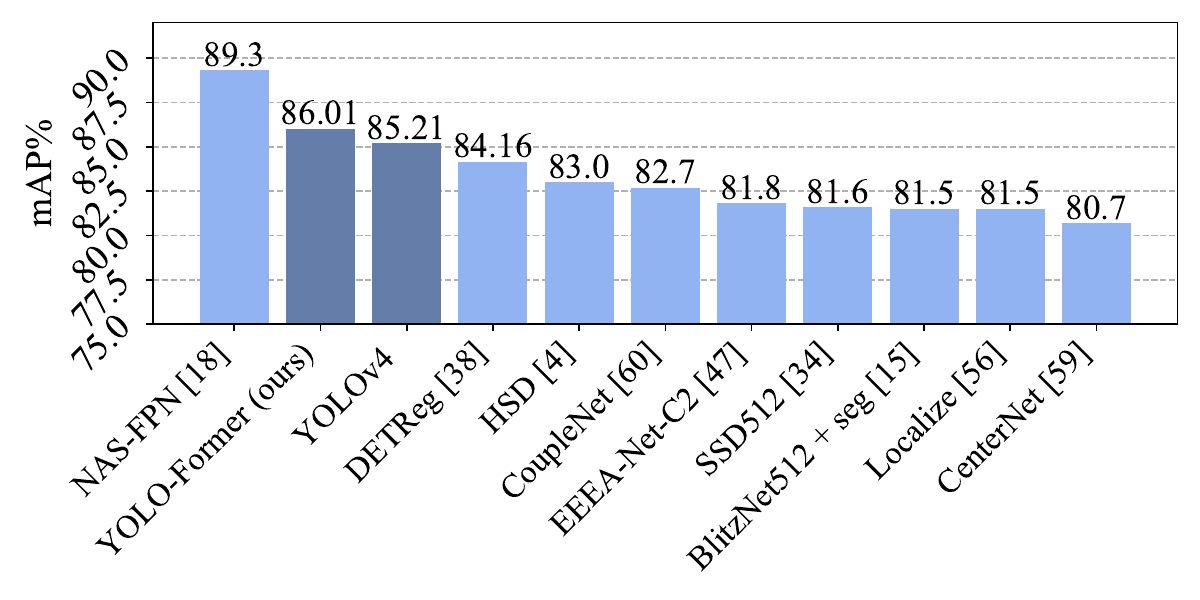}
	\centering
	\caption{The mean average precision (mAP) of YOLOv4 and YOLO-Former compared to the top-performing models on the Pascal VOC test set according to papers with code~\cite{paperswithcode}.}
	\label{fig:res2}
\end{figure}

Based on this study's most accurate YOLOv4 and YOLO-Former predictions, \reffig{fig:res1} illustrates each class's average precision (AP). In spite of YOLOv4's higher AP in five classes (potted plant, cow, bicycle, motorbike, and airplane), YOLO-Former's AP was equal to or greater in every other class. Besides that, the pattern is almost the same; both models have the most difficulties with the potted plant class, and airplane, bus, and horse are the most accessible classes to detect.
A comparison is made between YOLOv4 and YOLO-Former in terms of mAP against the most accurate models on the Pascal test set, depicted in \reffig{fig:res2}. The YOLO-Former and YOLOv4 outperforms all the previously evaluated works in the literature on Pascal VOC dataset such as DETReg \cite{maaz2022class}, HSD \cite{cao2019hierarchical}, CoupleNet \cite{zhu2017couplenet}, EEEA-Net-C2 \cite{termritthikun2021eeea}, SSD512 \cite{liu2016ssd}, BlitzNet512\cite{dvornik2017blitznet}, Localize \cite{zhang2020localize}, and CenterNet \cite{zhou2019objects}, except NAS-FPN \cite{ghiasi2021simple}. 


Since the YOLO network's ability to perform in real-time is an important characteristic, the prediction speed is worth examining. In this regard, 1000 random images are selected for the YOLOv4 and the YOLO-Former variants to make the prediction. The required time for each prediction is captured and averaged over all these test images. Testing is carried out on Tesla P100 at three different resolutions, 384x384, 416x416, and 512x512. The results are presented in \reftab{tab:res2}. All resolutions shows YOLOv4 to be faster than YOLO-Former models, but as the resolution increases, the gap decreases. Among the YOLO-Former variants, it is the fastest with the single-head attention module; with the combined multi-head and multi-branch (MHMB) attention module appears to be the slowest. 


\begin{table}
  \centering
  \begin{tabular}{|c|c|l|c|}
    \hline
    \centering
    & {Input Res.} & Model & FPS\\
     \hline
     &  & YOLOv4 & 16.74 \\
     & & YOLO-Former (SH) & 11.76\\
    1 &320x320 & YOLO-Former (MH) & 10.16 \\
     & & YOLO-Former (MB) & 9.76 \\
     & & YOLO-Former (MHMB) & 7.33\\
     \hline
       &  & YOLOv4 & 12.31 \\
	& & YOLO-Former (SH) & 10.85 \\
     2 & 416x416 & YOLO-Former (MH) & 9.09 \\
        & & YOLO-Former (MB) & 8.88 \\
	& & YOLO-Former (MHMB) & 7.08\\
     \hline
    & & YOLOv4 & 11.63 \\
     & & YOLO-Former (SH)&  10.27\\
    3 & 512x512 & YOLO-Former (MH) & 8.97 \\
     & & YOLO-Former (MB) & 8.02 \\
    & & YOLO-Former (MHMB) & 6.97\\
     \hline

  \end{tabular}
  
   \caption{The comparison of YOLOv4 and YOLO-Former prediction speed (frame per second) with various input resolutions.}
\label{tab:res2}
\end{table}

\section{Conclusion}
\label{sec:conclusion}
This study developed a real-time object detector called YOLO-Former, based on the idea of the transformer and YOLOv4. In order to accomplish this network, several convolutional self-attention modules were developed. The implemented model was trained with several settings and compared with YOLOv4 on the Pascal VOC test set. The RandAugment-trained YOLOv4 and YOLO-Former provided more accurate results than the AugMix-trained YOLOv4. Moreover, additional data was essential to achieve a desirable accuracy on the Pascal VOC test data. Based on the obtained results, it is concluded that through attention, YOLO-Former is capable of producing a higher accuracy at the cost of a drop in prediction speed. This descent becomes insignificant as the input resolution increases. Additionally, the single-head convolutional self-attention module offered the best accuracy-speed tradeoff. As such, the proposed YOLO-Former has provided a more accurate convolutional object detector compared to the similar existing methods on the Pascal VOC dataset.



\end{document}